\documentclass[review,sort&compress,3p]{elsarticle}

\usepackage{graphicx}
\usepackage{amssymb}
\usepackage{color}
\usepackage[colorlinks,
            linkcolor=red,
            anchorcolor=blue,
            citecolor=green
            ]{hyperref}
\usepackage{multirow}
\usepackage{booktabs}
\usepackage[titletoc,toc,title]{appendix}
\usepackage{amsmath,latexsym}
\usepackage{caption}
\usepackage{subfigure}
\usepackage{verbatim}
\usepackage{amsthm}
\usepackage{graphicx}
\usepackage{mathrsfs}
\usepackage{amsfonts}
\usepackage{algorithm2e}
\usepackage[noend]{algpseudocode}
\usepackage{float}
\makeatletter
\def\BState{\State\hskip-\ALG@thistlm}
\makeatother

\def\bx{\mathbf{x}}
\def\by{\mathbf{y}}

\def\bp{\mathbf{p}}
\def\bs{\mathbf{s}}

\def\M{\mathcal{M}}

\begin{document}

\begin{frontmatter}

\title{Weighted Nonlocal Total Variation in Image Processing}

\author[a]{Haohan Li}
\ead{hlibb@connect.ust.hk}
\author[b]{Zuoqiang Shi\corref{cor1}}
\ead{zqshi@tsinghua.edu.cn}
\author[a]{Xiao-Ping Wang}
\ead{mawang@ust.hk}

\
\cortext[cor1]{Corresponding author}
\address[a]{Department of Mathematics, The Hong Kong University of
                       Science and Technology, Hong Kong}
\address[b]{Department of Mathematical Sciences \& Yau Mathematical Sciences Center, Tsinghua University, Beijing, China}

\begin{abstract}
In this paper, a novel weighted nonlocal total variation (WNTV) method is proposed. 
Compared to the classical nonlocal total variation methods, our method modifies the energy functional to introduce a weight to balance 
between the labeled sets and unlabeled sets. With extensive numerical examples in semi-supervised clustering, image inpaiting and image colorization, 
we demonstrate that WNTV provides an effective and efficient method in many image processing and machine learning problems.
\end{abstract}

%\begin{keyword}

%\end{keyword}

\end{frontmatter}

\section{Introduction\label{sec1}}

Interpolation on point cloud in high dimensional space is a fundamental problem in many machine learning and image processing applications.
It can be formulated as follows.
 Let
$P = \{\bp_1, \cdots, \bp_n\}$ be a set of points in $\mathbb{R}^d$ and $S =\{\bs_1, \cdots, \bs_m\}$
be a subset of $P$. Let $u$ be a function on the point set $P$ and the value of $u$ on $S\subset P$ is given
as a function $g$ over $S$. 
%i.e. $u(\bs)=g(\bs),\;\forall \bs\in S$.
The goal of the interpolation is to find the function $u$ on $P$ with the given values on $S$.

Since the point set $P$ is unstructured in high dimensional space, traditional interpolation methods do not apply. In recent years, manifold learning 
has been demonstrated to be effective and attract more and more
attentions. One basic assumption in manifold learning is that the point cloud $P$ samples a low dimensional smooth manifold, $\mathcal{M}$,
 embedded in $\mathbb{R}^d$. 
Another assumption is that the interpolation function $u$ is a smooth function in $\mathcal{M}$. Based on these two assumptions, one popular approach is to solve $u$ by minimizing the 
$L_2$ norm of its gradient in $\mathcal{M}$. This gives us an optimization problem to solve:
\begin{equation}
\label{eq:ob-l2}
\min_{u} \|\nabla_\mathcal{M}u\|_2,\quad \text{subject to:}\quad u(\bx)=g(\bx),\quad \bx\in S,
\end{equation}
with $$\|\nabla_\mathcal{M}u\|_2=\left(\int_\mathcal{M} |\nabla_\M u(\bx)|^2 d\bx\right)^{1/2}.$$
Usually, $\nabla_\M u$ is approximated by nonlocal gradient
\begin{equation}
\label{eq:nlgrad}
D_{\by} u(\bx)=\sqrt{w(\bx,\by)}(u(y)-u(\bx)).
\end{equation}
Then, the discrete version of \eqref{eq:ob-l2} is 
 \begin{equation}
\label{eq:ob-gl}
\min \sum_{\bx,\by\in P} w(\bx,\by)(u(\bx)-u(\by))^2,\quad  \text{subject to:}\quad u(\bx)=g(\bx),\quad \bx\in S,
\end{equation}
from which, we can derive a linear system to solve $u$ on point cloud $P$.
This is just the well known nonlocal Laplacian which is widely used in nonlocal methods of image processing \cite{BCM05,BCM06,GO07,GO08}. 
It is also called graph Laplacian in graph and machine learning literature ~\cite{Chung:1997,ZhuGL03}. Recently, it was found that, when the sample rate is low, i.e. $|S|/|P|\ll1$,
graph Laplacian method fails to give a smooth interpolation \cite{Shi:harmonic, Shi2017}. Continuous interpolation can be obtained by using point integral method \cite{Shi:harmonic} or 
weighted nonlocal Laplacian \cite{Shi2017}.

In many problems, such as data classification or image segmentation, minimizing the total variation seems to be a better way to compute the interpolation function, since it 
prefers piecewise constant function in total variation minimization. This observation motives  another optimization problem:   
\begin{equation}
\label{eq:ob-tv}
\min \|u\|_{TV_\M},\quad \text{subject to:}\quad u(\bx)=g(\bx),\quad \bx\in S,
\end{equation}
with $$\|u\|_{TV_\M}=\int_\mathcal{M} |\nabla_\M u(\bx)| d\bx.$$

Total variation model has been studied extensively in image processing since it was first proposed by Rudin, Osher and Fatemi(ROF) in \cite{rudin1992nonlinear}. 
% The ROF denoising model is designed to solve the following functional:
% \begin{align*}
% \min_u ||u||_{TV}+\frac{\mu}{2}||u-f||^2_{L^2}\label{TV}
% \end{align*}
% where $f$ is an observed image and $u$ is the true image. $||u||_{TV} = \int |\bigtriangledown u(x)|dx$ is the total variation of function $u$. 
It is well known that total variation has the advantage of preserving
edges, which is always preferable because edges are significant features in the image, and
usually indicate boudaries of objects. Despite its good performance of restoring "cartoon"
part of the image, TV based methods fail to achieve satisfactory results when texture, or
repetative structures, are present in the image.
To address this problem, Buades et al proposed a nonlocal means
method based on patch distances for image denoising \cite{BCM05}. Later, Gilboa and Osher \cite{GO07,GO08}
formalized a systematic framework, include nonlocal total variation model, for nonlocal image processing. 

Using nonlocal gradient to approximate the total variation, we can write down the discrete version of (\ref{eq:ob-tv}),
\begin{equation}
\label{eq:ob-tv-dis}
\min \sum_{\bx \in P}\left(\sum_{\by\in P} w(\bx,\by)(u(\bx)-u(\by))^2\right)^{1/2},\quad \text{subject to:}\quad u(\bx)=g(\bx),\quad \bx\in S.
\end{equation}
This problem can be solved efficiently by split Bregman iteration \cite{osher2005iterative,doi:10.1137/080725891}. 
However, it was reported that \cite{LDMM}, when the sample rate is low, above nonlocal TV model has the same defect as that in graph Laplacian approach \eqref{eq:ob-gl}.
The interpolation obtained by solving above optimization problem is not continuous on the sample points. 
% In those papers, the authors propose a Split Bregman method, which can solve various l1-regularized problems in image processing and compressed sensing efficientlly. Hence, Bregman iteration has already been applied to a broad class of problems in the previous fields that involves l1-regularization problems. For example, Cai at. al. derived an algorithm for compressed sensing problems via Bregman iteration in \cite{cai2009linearized}. Other applications to compressed sensing could be found in \cite{osher2011fast,yin2008bregman}. Problems in medical imaging can also be solved using Bregman techniques \cite{chang2006mr}. We will introduce more about the method in next . 

In this paper, inspired by weighted nonlocal laplacian method proposed in \cite{Shi2017}, we propose a weighted nonlocal TV method (WNTV) to fix this discontinuous issue. 
The idea is to introduce a weight related to the sample rate to balance the labeled terms and unlabeled terms. More specifically, we modify model \eqref{eq:ob-tv-dis} a little bit 
by introducing a weight,
\begin{align*}
\min_u \sum_{x\in V\backslash S}\left(\sum_{y\in V} \omega(x, y)(u(x)-u(y))^2\right)^{1/2} + \frac{|V|}{|S|}\sum_{x\in S}\left(\sum_{y\in V} \omega(x, y)(u(x)-u(y))^2\right)^{1/2},\label{WNTV}
\end{align*}
This optimization problem also can be solved by split Bregman iteration. Based on our experience, the convergence is even faster than the split Bregman iteration in the original
nonlocal total variation model \eqref{eq:ob-tv-dis}. Using extensive examples in image inpaiting, semi-supervised learning, image colorization, we demonstrate that the weighted 
nonlocal total variation model has very good performance. It provides an effective and efficient method for many image processing and machine learning problem. 
 
% It is well known the nonlocal methods are widely used in a broad class of applications. In particular TV based semi-supervised learning models are good at restoring cartoon like images. However, it was observed that the performance of the proposed method decreases when one increases the number of unlabeled data. That is when the sample rate is low, the TV based method will fail in these applications. Similar phenomenon also arise in graph laplacian based methods. In order to solve this problem, we proposed this weighted nonlocal TV method.
 
The rest of the paper is organized as follows. In section \ref{sec1}, we review the interpolation problem on point cloud, which is typically hard to solve by traditional interpolation method. Then the weighted nonlocal TV method (WNTV) is introduced in section \ref{sec2}. We apply the split Bregman iteration algorithm to our method, which is a well-known algorithm to solve a very broad class of L1-regularization problems. Numerical experiments including semi-supervised clustering, image inpainting and image colorization are shown in section \ref{sec3}, \ref{sec4} and \ref{sec5} respectively. Here we compared our results to those obtained using graph Laplacian, nonlocal TV and weighted nonlocal Laplacian. Conclusions are made in the section \ref{sec6}.

\section{Weighted Nonlocal TV\label{sec2}}

As introduced at the beginning of the introduction, we consider an interpolation problem in a high dimentional point cloud.
 Let
$V = \{\bp_1, \cdots, \bp_n\}$ be a set of points in $\mathbb{R}^d$ and $S =\{\bs_1, \cdots, \bs_m\}$
be a subset of $V$.  $u$ is a function on $V$ and $u(\bs)=g(\bs),\;\forall \bs\in S$ with given $g$.
We assume that $V$ samples a smooth manifold $\M$ embedded in $\mathbb{R}^d$ and we want to minimize the total variation of $u$ on $\M$ to solve $u$ on the whole poing cloud $V$.
This idea gives an optimization problem in continuous version:
\begin{equation}
\label{eq:ob-nltv}
\min_u \int_\mathcal{M} |\nabla_\M u(\bx)| d\bx,\quad \text{subject to:}\quad u(\bx)=g(\bx),\quad \bx\in S,
\end{equation}
Using the nonlocal gradient in \eqref{eq:nlgrad} to approximate the gradient $\nabla_\M u$, we have a discrete optimization problem
% Let $u$ be a function on the graph vertices $V$. Given a pair of points $(x, y)\in V\times V$, then the nonlocal gradient is defined to be:
% \begin{align*}
% \bigtriangledown_{\omega}u(x, y) = \sqrt{\omega(x, y)}(u(x)-u(y)).
% \end{align*}
% The norm of interest here is the following nonlocal total variation:
% \begin{align}
% ||\bigtriangledown_{\omega}u(x, y)||_{L_1} = \sum_x\left(\sum_y \omega(x, y)(u(x)-u(y))^2\right)^{1/2}
% \end{align}
\begin{equation}
\label{eq:ob-nltv-dis}
\min \sum_{\bx \in V}\left(\sum_{\by\in V} w(\bx,\by)(u(\bx)-u(\by))^2\right)^{1/2},\quad \text{subject to:}\quad u(\bx)=g(\bx),\quad \bx\in S,
\end{equation}

%\begin{figure}[H]
%\centering
  % \subfigure[]{ \includegraphics[width=0.4\textwidth]{images/barb/NTVbarb.jpg}}
  % \subfigure[]{ \includegraphics[width=0.4\textwidth]{images/bf/NTVbf.jpg}}
 %  \caption{\label{inpaint-intro} Recovered image by Nonlocal TV from 10\% random samples.}
%\end{figure}

Inspired by the weighted nonlocal Laplacian method proposed by Shi et. al. in \cite{Shi2017}, we actually modify the above functional to add weight to balance the energy between labeled points and unlabeled sets:
\begin{align}
\min_u \sum_{\bx\in V\backslash S}\left(\sum_{\by\in V} \omega(\bx, \by)(u(\bx)-u(\by))^2\right)^{1/2} 
+ \frac{|V|}{|S|}\sum_{\bx\in S}\left(\sum_{\by\in V} \omega(\bx, \by)(u(\bx)-u(\by))^2\right)^{1/2},\label{WNTV}
\end{align}
with the constraint 
\begin{align}
u(\bx) = g(\bx), \bx\in S.
\end{align}
where $S$ is a subset of the vertices set $V$, and $|V|, |S|$ are the number of points in sets $V$ and $S$, respectively.
The idea is that
when the sample rate is low, the summation over the unlabeled set overwhelms the summation over the labeled set such that the continuity
on the labeled set is sacrificed. To maintain the continuity of the interpolation on the labeled points, we introduce a weight to balance the labeled term and the unlabeled term.

The weighted nonlocal total variation model (WNTV) \eqref{WNTV} can be solved by split Bregman iteration \cite{doi:10.1137/080725891}.
To simplify the notation, we introduce an operator as follows,
\begin{align*}
D_{NG}u(\bx, \by) = \left\{ \begin{array}{lll}
    \sqrt{\omega(\bx, \by)} (u(\bx) - u(\by)), &  & \textbf{if} \ \ \ \bx \in V \backslash S,\\
    &  & \\
    \frac{|V|}{|S|} \sqrt{\omega(\bx, \by)} (u(\bx) - u(\by)), &  & \textbf{if} \ \ \ \bx \in
    S.
  \end{array} \right. \label{NG}
\end{align*}
%To apply the Bregman splitting, we replace $D_{NG}u(x, y)$ by $D(x, y)$, which yields the following constrained problem,
With above operator, WNTV model \eqref{WNTV} can be rewritten as
\begin{align}
\min_{u, D} \sum_{x\in V}\left(\sum_{y\in V}|D(\bx, \by)|^2\right)^{1/2},\quad \text{subject to:}\ \ \  D(\bx,\by) = D_{NG}u(\bx,\by).
\end{align}
with the constraint 
\begin{align*}
u(\bx) = g(\bx), x\in S.
\end{align*}
We then use Bregman iteration to enforce the constraint $D(\bx,\by) = D_{NG}u(\bx,\by)$ to get a two-step iteration,
\begin{align}
\label{SB}
(u^{k+1},D^{k+1})=\arg&\min_{u, D} \sum_{x\in V}\left(\sum_{y\in V}|D(x, y)|^2\right)^{1/2} + \frac{\lambda}{2}\sum_{x\in V}\sum_{y\in V}\left(D(x, y)-D_{NG}u(x, y) - Q^k(x, y)\right)^{2}, \\
&\text{subject to:} \quad  u(\bx) = g(\bx), x\in S.\nonumber\\
Q^{k+1} = Q^k &+ (D_{NG}u^{k+1} - D^{k+1}).\label{ite:Q}
\end{align}
where $\lambda$ is a positive parameter.

In above iteration, \eqref{ite:Q} is easy to solve. 
To solve the minimization problem (\ref{SB}), we use the idea in the split Bregman iteration to solve $u$ and $D$ alternatively.% note that the variables $u$ and $D$ are now decoupled, we can actually decompose it into two subproblems and alternately minimize the subproblems. More precisely,
\begin{align}
%\left\{ \begin{array}{lll}
    u^{k+1} =& \arg\min\limits_{u} ||D^{k}-D_{NG}u - Q^{k}||^2_2,\quad \text{subject to:} \quad  u(\bx) = g(\bx), x\in S. \\
    D^{k+1} =& \arg\min\limits_{D} \|D\|_1 + \frac{\lambda}{2}||D-D_{NG}u^{k+1} - Q^{k}||^2_2. \label{OP2}\\
    Q^{k+1} = &Q^k + (D_{NG}u^{k+1} - D^{k+1}).
 % \end{array} \right. \label{OP}
\end{align}
where 
$$\|D\|_1=\sum_{\bx\in V}\left(\sum_{\by\in V}|D(\bx, \by)|^2\right)^{1/2}.$$
 The first step is a standard least-squares problem. It is staightforward to see that $u^{k+1}$ satisfies a linear system,
\begin{align}
&\sum_{\by\in V\backslash S} (\omega(\bx,\by) + \omega(\by,\bx))(u(\bx) - u(\by)) + 
\sum_{\by\in S} \left(\omega(\bx,\by) + \left(\frac{|V|}{|S|}\right)^2\omega(\by, \bx)\right)(u(\bx) - u(\by)) 
\nonumber\\
&- \sum_{\by\in V\backslash S}\left((D^k(\bx,\by)-Q^k(\bx,\by))\sqrt{\omega(\bx, \by)} - (D^k(\by,\bx)-Q^k(\by,\bx))\sqrt{\omega(\by,\bx)}\right) 
\nonumber\\
&- \sum_{y\in S}\left((D^k(\bx,\by)-Q^k(\bx,\by))\sqrt{\omega(\bx,\by)} - \frac{|V|}{|S|}(D^k(y,x)-Q^k(y,x))\sqrt{\omega(\by,\bx)}\right)  = 0, \quad \bx\in V\backslash S, \label{Sou}
\end{align}
with the constraint 
\begin{align}
\label{eq:label}
u(\bx)  =  g(\bx), \quad \bx\in S.
\end{align}
The linear system \eqref{Sou}-\eqref{eq:label} looks like complicated. Its coefficient matrix is sparse, symmetric and postive definite which can be solved efficiently by conjugate 
gradient method. 

The minimizer of the optimization problem (\ref{OP2}) 
can be explicitly computed using shrinkage operators \cite{doi:10.1137/080725891}. Notice that this problem is decoupled 
in terms of $\bx$, i.e. $D_x=D(x,:)$ actually solves a subproblem,
\begin{align*}
\min_{D_\bx}\quad |D_\bx| + \frac{\lambda}{2}||D_\bx-D_{NG\bx} u^{k+1} - Q_\bx^{k}||^2_2, %\ \ \ \text{for} \ \ \ \bx\in V
\end{align*}
where 
$$|D_\bx|=\left(\sum_{\by\in V}|D(\bx, \by)|^2\right)^{1/2},$$
and $D_{NG\bx}u^{k+1}=D_{NG}u^{k+1}(\bx,:), \;Q_\bx^k=Q^k(\bx,:)$.

%and the shrinkage formula for the above optimal problem is given by,
It is well known that solution of above optimization problem can be given by soft shrinkage. 
\begin{align*}
D^{k+1}_\bx = \text{shrink}(D_{NG\bx} u^{k+1} + Q_\bx^{k},1/\lambda)
\end{align*}
where
\begin{align*}
\text{shrink}(\mathbf{z},\gamma) = \frac{\mathbf{z}}{||\mathbf{z}||_2}\max(||\mathbf{z}||_2-\gamma,0)
\end{align*}

%Using the above iterative minimization, the optimization problem can be solved with the following formulations,
Summarizing above discussion, we get an iterative algorithm to solve weighted nonlocal total variation model,
\begin{algorithm}%[H]
\floatname{algorithm}{Algorithm}
\caption{Algorithm for WNTV}
\label{SBWTV}
%\begin{itemize}
1. Solve \eqref{Sou}-\eqref{eq:label} to get $u^{k+1}$.

2. Compute $D^{k+1}$ by 
  \begin{align*}
    D^{k+1}(\bx,\by)
    = \frac{\bar{D}(\bx,\by)}
{\left(\displaystyle \sum_{\by\in V} |\bar{D}(\bx,\by)|^2\right)^{1/2}}
\max\left(\sqrt{\sum_{\by\in V} |\bar{D}(\bx,\by)|^2}-\frac{1}{\lambda},0\right)
  \end{align*}
with $\bar{D}(\bx,\by)=\sqrt{\omega(\bx,\by)}(u^{k+1}(\bx)-u^{k+1}(\by))+Q^k(\bx,\by)$.

3. Update $Q$ by
$$Q^{k+1}_{i j} = Q^k_{i j} + ((D_{NG} u^{k+1})_{i j} - D^{k+1}_{i j}).$$
\vspace{3mm}
%\end{itemize}
\end{algorithm}

\section{Semi-supervised Clustering\label{sec3}}
\begin{figure}[h]
\centering
        \includegraphics[width=0.7\textwidth, height=0.4\textwidth]{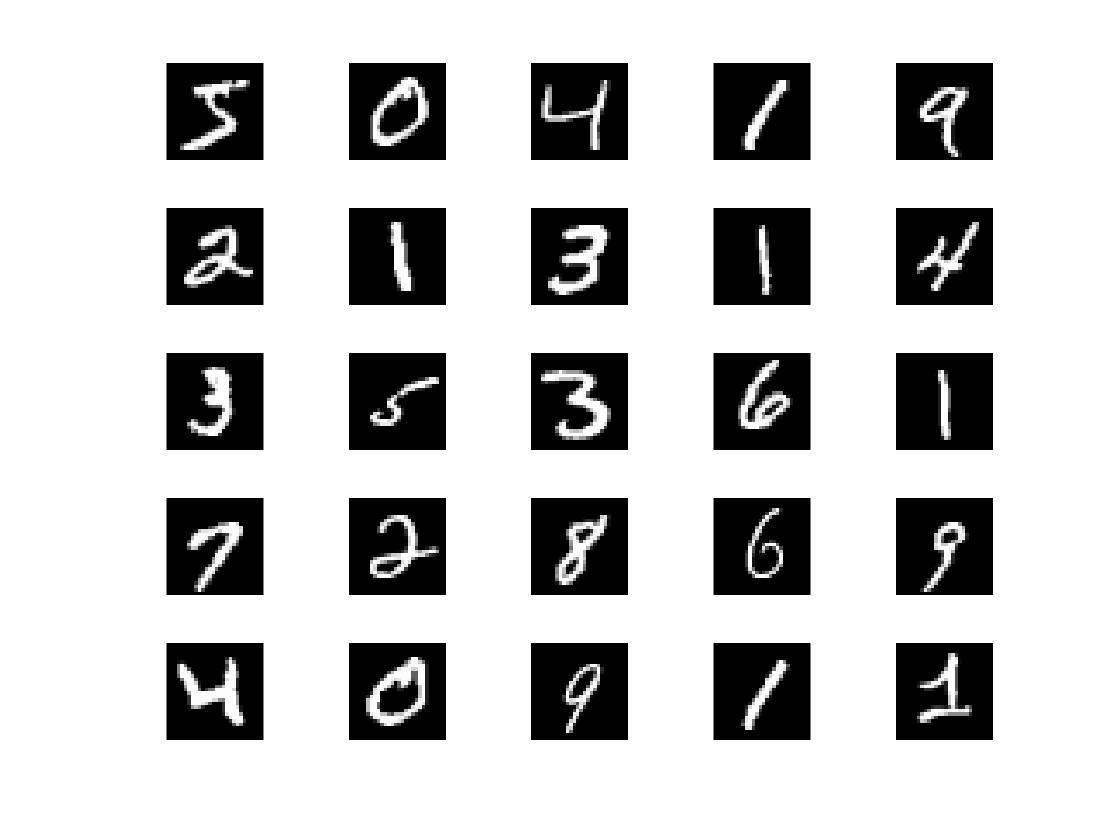}
  \caption{\label{fig:mnistexample}Some examples in the MNIST handwritten digits dataset}
\end{figure}

In this section, we test WNTV in a semi-supervised clustering problem on the famous MNIST data set \cite{mnist}. %It is available at http://yann.lecun.com/exdb/mnist/. 
The MNIST database consists of 70,000 grayscale 28$\times$28 pixel images of handwritten digits, see Fig. \ref{fig:mnistexample}, 
which is divided into a training set of 60,000 examples, and a test set of 10,000 examples. The images include digits from 0 to 9, which can be viewed as 10 classes segmentation.

From geometrical point of view, 70,000 28$\times$28 images form a point cloud $V$ in 784-dimension Euclidean space. 
In the tests, we randomly select a small subset $S\subset V$ to label, 
\begin{align*}
S = \bigcup\limits^l_i S_i,
\end{align*}
where $S_i$ is a subset of $S$ with label $i$. Our task here is to label the rest of unlabeled images.
The algorithm we used is summarized in Algorithm \ref{ssl}.

\begin{algorithm}[H]
\caption{Semi-Supervised Learning}\label{ssl}
\KwData{A set of points $V$ with a small subset labeled $S=\bigcup\limits^l_i S_i$}
 \KwResult{ labels of the whole points set $V$ }
1. Compute the corresponding weight function $\omega(\bx,\by)$ for $\bx,\by\in V$\;
  \For{$i=0:9$}{
%{
%  2. initialize $u_i$ such that 
%  \begin{align*}
% u_i(x) = \left\{ \begin{array}{lll}
%     1, &  & \textbf{if} \ \ \ x \in S_i,\\
%     &  & \\
%     0, &  & \textbf{otherwise}.
%   \end{array} \right.
% \end{align*}
%  }
 2. Compute $u_i$ by WNTV using Algorithm \ref{SBWTV} with the constraint
$$u_i(\bx)=1, \;\bx \in S_i,\quad u_i(\bx)=0, \; \bx\in S\backslash S_i.$$
}
 3. Label $x\in V\setminus S$ as $k$ when $k=\arg \max\limits_{1\leq i\leq l}u_i(x)$
\vspace{5mm}
\end{algorithm}
\vspace{0.5cm}

In our experiment of MNIST dataset, the weight function $\omega(\bx,\by)$ 
is then constructed using the Gaussian,
\begin{align*}
\omega(\bx, \by) = \exp\left(-\frac{\|\bx-\by\|^2}{\sigma(\bx)^2}\right),
\end{align*}
where $\|\cdot\|$ denotes the Euclidean distance, $\sigma(\bx)$ is the distance between $\bx$ and its 10th nearest neighbor. The weight $\omega(\bx,\by)$ is made sparse by setting $\omega(\bx,\by)$ equal to zero if point $\by$ is not among the 20th closest points to point $\bx$.

\begin{table}
\centering
\begin{tabular}{l||ccc}
\hline %
Methods 
                             & 700/70000 & 100/70000 & 50/70000\\\hline %
 WNTV               & 94.08 & 89.86 & 78.35\\\cline{1 -4}
 Nonlcal TV             & 93.78 & 32.55 & 28.00\\\hline
WNLL                  & 93.25 & 87.84 & 73.60\\\cline{1 -4}
   GL            & 93.15 & 35.17 & 20.09\\\hline
\end{tabular}
\caption{\label{sslresult} Rate of correct classification in percentage for MNIST dataset}
\end{table}

From the result of table (\ref{sslresult}), we can see that with high label rate (700/70000), all four methods give good classification. 
Nevertheless, as the label rate is reduced (100/70000, 50/70000), graph Laplacian and nonlocal TV both fail. The results given by WNTV and WNLL still have reasonable accuracy. 
WNTV is slightly better than WNLL in our tests. 

\section{Image Inpainting\label{sec4}}
The problem of fitting the missing pixels of a corrupted image is always of interest in image processing. This problem can be formulated as an interpolation problem 
on point cloud by considering patches of the image. Consider a discrete image $f\in \mathbb{R}^{m\times n}$, around each pixel $(i,j)$, we define a patch $p_{i j}(f)$ that is $s_1 \times s_2$ collection of pixels of image $f$. %The feature vectors associated with each pixel $(i,j)$ is constructed from the patch $p_{i j}(I)$. 
The collection of all patches is defined to be the patch set $\mathscr{P}(f)$ \cite{osher2016low},
\begin{align*}
\mathscr{P}(f) = \{ p_{i j}(f):(i,j)\in \{1,2,...,m\}\times \{1,2,...,n\}\}
\end{align*}
Here $\mathscr{P}(f)$ forms a point set $V$. %The proper choice of patch size $s_1 \times s_2$ and also features are application dependent. 

Then the image can be viewed as a function $u$ on the point cloud $\mathscr{P}(f)$. $u$ is defined to be the intensity of the central pixel of the patch,
\begin{align*}
u(p_{i j}(f)) = f(i,j),
\end{align*}
where $f(i,j)$ is the intensity of pixel $(i,j)$. 

Now, given subsample of the image, the problem of image inpainting is to fit the missing value of $u$ on the patch set $\mathscr{P}(f)$. However, this problem is actually
more difficult than the interpolation, since the patches is also unknown. In the image inpainting, we also need to recover the point cloud in addition to the interpolation function.
We achieve this by a simple iterative scheme. First, we fill in the missing pixels by random number to get a complete image. For this complete (quality is bad) image, we construct
point cloud by extracting patches. On this point cloud, we run WNTV to compute an interpolation function. From this interpolation function, we can construct an image. Then the 
patch set is updated from this new image. By repeating this process until convergence, we get the restoration of the image.  
% Because the original image $f$ is supposed to be unknown, we need to update the patch set $\mathscr{P}(f)$ constructed from the recovered image every iterative step. First, we subsample an image randomly to get a subset $S$ of $V$. Here $V$ is the set of all pixels on image, and $f_V$ is the original whole image while $f_S$ is the image with only pixels in $S$ shown. The value of $u$ is given in $S\subset V$, not known outside $S$. Our task here is to calculate the value of $u$ in $V\setminus S$.
We summarize this ideas in algorithm (\ref{imrecal}).

\vspace{0.5cm}
\begin{algorithm}[H]
\caption{Image Inpainting}\label{imrecal}
\KwData{A subsampled image}
 \KwResult{ Recovered image $u$ }
 initialize $u^0$ such that $u^0_S = f_S$ and $D^0,Q^0 = 0$\;
 \While{not converge}{
  1. Construct patch set $\mathscr{P}(u^n)$ from the current recovered image $u^n$ at step $n$\;
  2. Compute the corresponding weight function $\omega^n(x,y)$ for $x,y\in \mathscr{P}(u^n)$\;
  3. Compute $u^{n+1}$ by solving system (\ref{SBWTV}),then update image correspondingly\;
  4. \textbf{goto} step 1\;
  }
\end{algorithm}
\vspace{0.5cm}

\subsection{Grayscale image inpainting}
We first apply the algorithm to grayscale images. In this case, we also use Gaussian weight,
\begin{align*}
\omega(\bx,\by) = \exp\left(-\frac{\|\bx-\by\|^2}{\sigma(\bx)^2}\right)
\end{align*} 
where $\|\bx-\by\|^2$ is the Euclidean distance between patches $\bx$ and $\by$. $\sigma(\bx)$ is the distance between $\bx$ and its 20th nearest neighbor. 
The weight $\omega(\bx,\by)$ is made sparse by setting $\omega(\bx,\by)$ equal to zero if point $\by$ is not among the 50th closest points to point $\bx$. 
For each pixel, we assign a 11$\times$11 patch around it consisting of intensity values of pixels in the patch. 
In order to accelerate the speed of convergence, we use the semi-local patch by adding the local coordinate to the end of the patches,
\begin{align*}
p_{i j}(I) = [p_{i j}, \lambda_1 i, \lambda_2 j]
\end{align*}
where
\begin{align*}
\lambda_1 = \frac{3||f_S||_\infty}{m}, \ \ \ \lambda_2 = \frac{3||f_S||_\infty}{n}.
\end{align*}

An approximate nearest neighbor algorithm (ANN) is used to obtain nearest neighbors. We use the Peak Signal-to-Noise Ratio (PSNR) to measure the quality of restored images,
\begin{align*}
\text{PSNR}(u,u_{gt}) = -20\log_{10}(\|u-u_{gt}\|/255)
\end{align*}
where $u$ and $u_{gt}$ are the restored image and the original image respectively.

\begin{figure}%[H]
\centering
   \subfigure[Original Image.\label{fig:originalimagebarb}]  {\includegraphics[width=4cm]{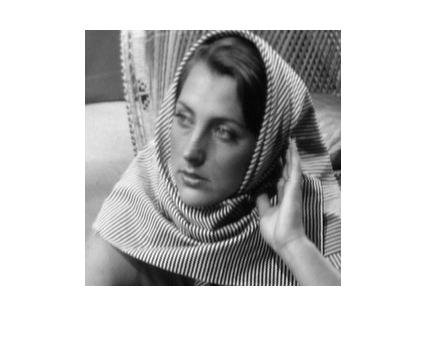}}
   \subfigure[10\% Subsample.\label{fig:sampleimagebarb}]{\includegraphics[width=4cm]{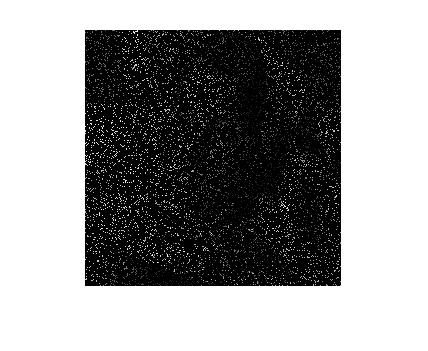}}
   \subfigure[GL (23.33dB)\label{fig:GLbarb}]{ \includegraphics[width=4cm]{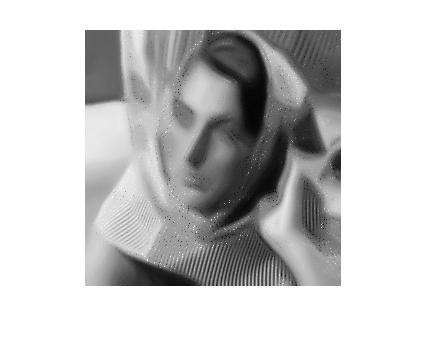}}\\
   \subfigure[NTV (22.85dB).\label{fig:NTVbarb}]{ \includegraphics[width=4cm]{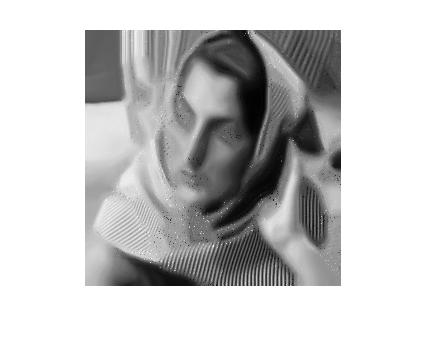}}
   \subfigure[WNLL (25.35dB).\label{fig:WGLbarb}]{ \includegraphics[width=4cm]{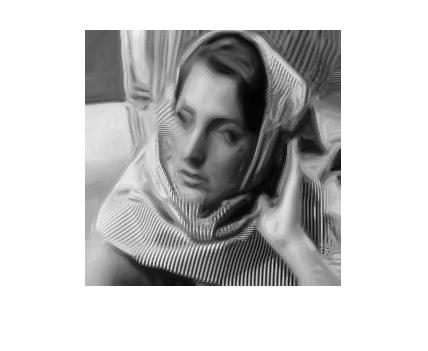}}
   \subfigure[WNTV (25.52dB).\label{fig:WNTVbarb}]{ \includegraphics[width=4cm]{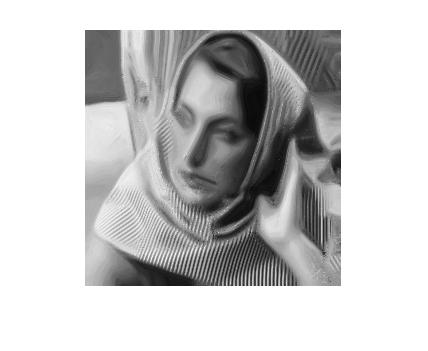}}
   \caption{\label{barbreult}Results of Graph Laplacian (GL), nonlocal TV (NTV), weighted nonlocal Laplacian (WNLL) and weighted nonlocal TV (WNTV) in image of Barbara. % (a): Given image; (b): Subsampled image 10\% sampling; (c): Result of Graph Laplacian, PSNR = 23.33; (d): Result of nonlocal TV, PSNR = 22.85; (e): Result of weighted nonlocal Graph Laplacian, PSNR = 25.35; (f): Result of weighted nonlocal TV, PSNR = 25.52.
   }
\end{figure}

\begin{figure}%[H]
\centering
   \subfigure[Original Image.\label{fig:originalimagebf}]  {\includegraphics[width=3.5cm]{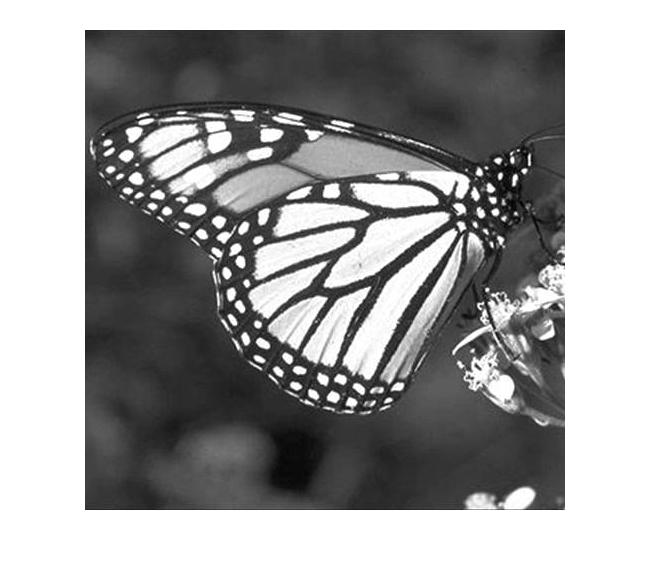}}
   \subfigure[10\% Subsample.\label{fig:sampleimagebf}]{\includegraphics[width=3.5cm]{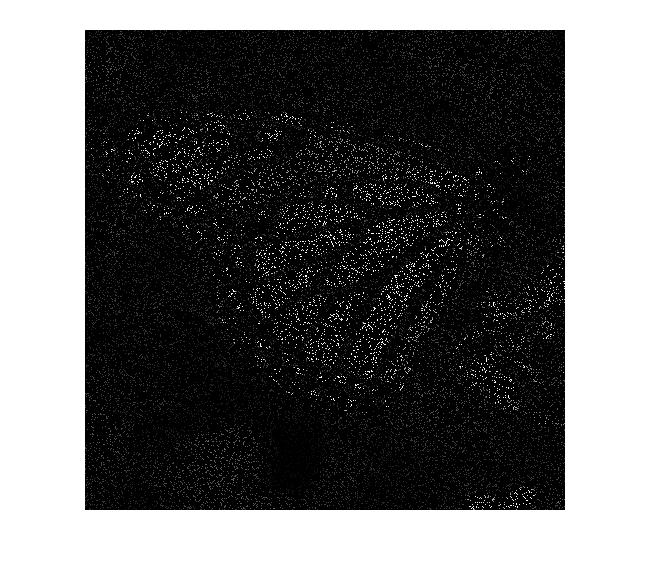}}
   \subfigure[GL (18.03dB).\label{fig:GLbf}]{ \includegraphics[width=3.5cm]{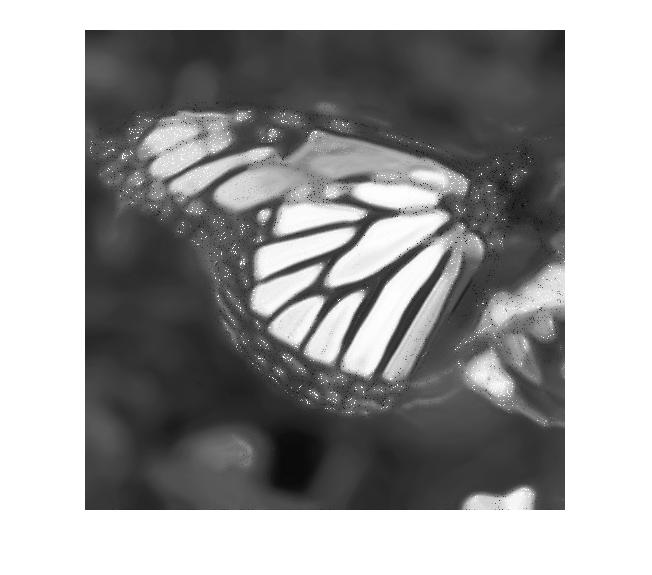}}\\
   \subfigure[NTV (17.89dB).\label{fig:NTVbf}]{ \includegraphics[width=3.5cm]{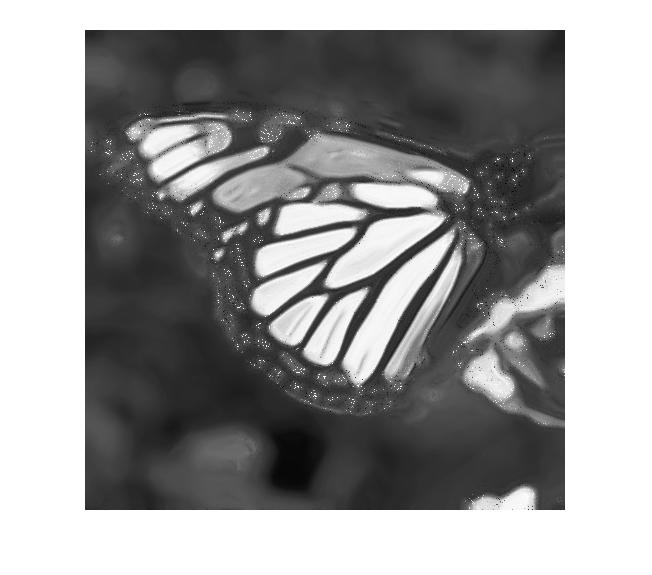}}
   \subfigure[WNLL (20.28dB).\label{fig:WGLbf}]{ \includegraphics[width=3.5cm]{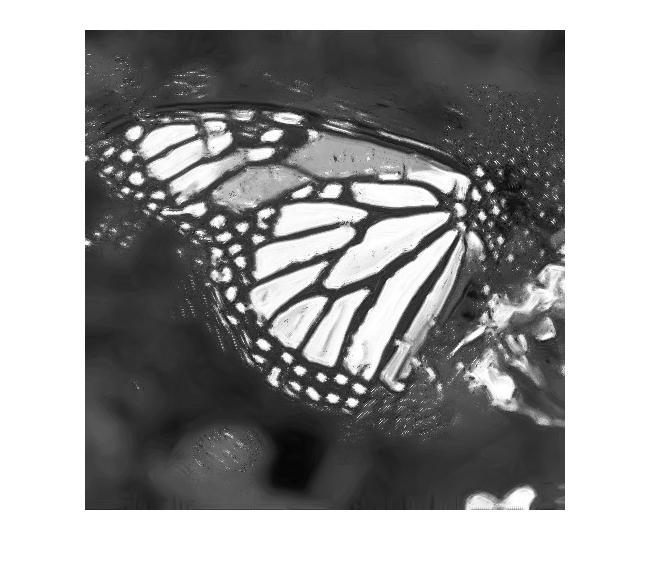}}
   \subfigure[WNTV (20.46dB).\label{fig:WNTVbf}]{ \includegraphics[width=3.5cm]{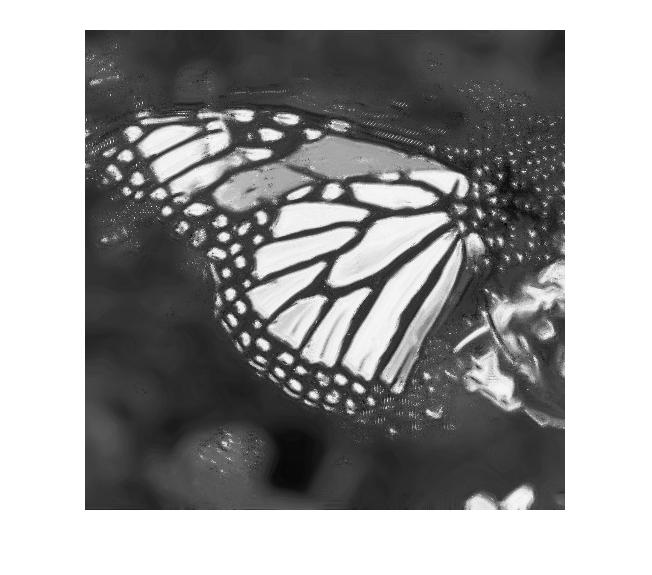}}
   \caption{\label{bfreult}Results of Graph Laplacian (GL), nonlocal TV (NTV), weighted nonlocal Laplacian (WNLL) and weighted nonlocal TV (WNTV) in the butterfly image. % (a): Given image; (b): Subsampled image 10\% sampling; (c): Result of Graph Laplacian, PSNR = 18.03; (d): Result of nonlocal TV, PSNR = 17.89; (e): Result of weighted Graph Laplacian, PSNR = 20.28; (f): Result of weighted nonlocal TV, PSNR = 20.46.
   }
\end{figure}

\begin{figure}%[H]
\centering
   \subfigure[Original Image.\label{fig:originalimagepp}]{\includegraphics[width=4cm]{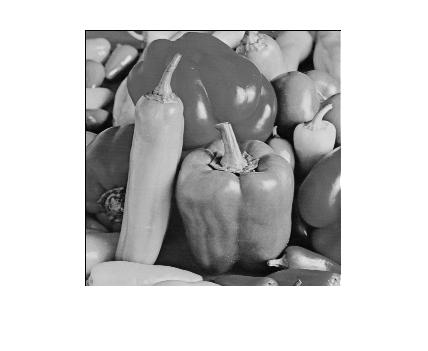}}
   \subfigure[Subsampled Image.\label{fig:sampleimagepp}]{\includegraphics[width=4cm]{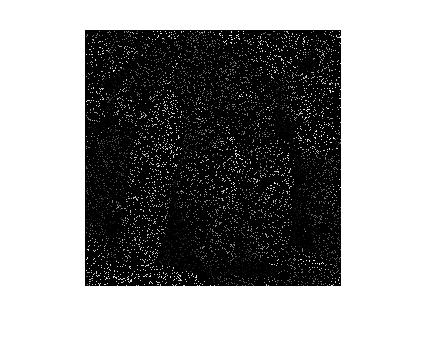}}
   \subfigure[GL (20.54dB).\label{fig:GLpp}]{\includegraphics[width=4cm]{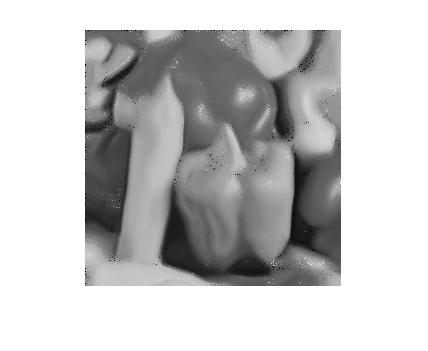}}\\
   \subfigure[NTV (20.93dB).\label{fig:NTVpp}]{\includegraphics[width=4cm]{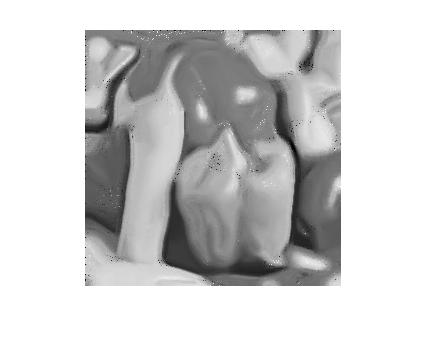}}
   \subfigure[WNLL (22.80dB).\label{fig:WGLpp}]{\includegraphics[width=4cm]{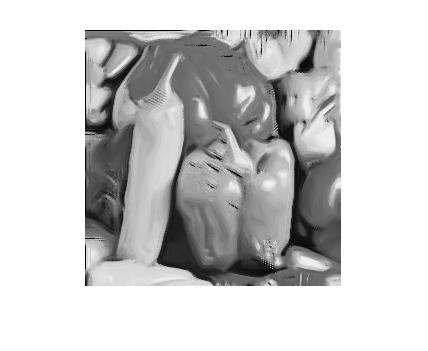}}
   \subfigure[WNTV (23.03dB).\label{fig:WNTVpp}]{\includegraphics[width=4cm]{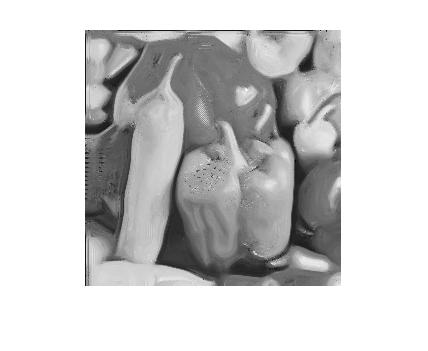}}
   \caption{\label{ppreult}Results of Graph Laplacian (GL), nonlocal TV (NTV), weighted nonlocal Laplacian (WNLL) and weighted nonlocal TV (WNTV) in the pepper image. % (a): Given image; (b): Subsampled image 10\% sampling; (c): Result of Graph Laplacian, PSNR = 20.54; (d): Result of nonlocal TV, PSNR = 20.93; (e): Result of weighted Graph Laplacian, PSNR = 22.80; (f): Result of weighted nonlocal TV, PSNR = 23.03.
   }
\end{figure}

The results are displayed in Fig. \ref{barbreult}, \ref{bfreult} and \ref{ppreult}. For each image, we fix the number of iterations to be 10. 
As we can see, WNTV and WNLL performs much better than classical nonlocal TV method and graph Laplacian. The results of WNLL are comparable to proposed WNTV. As expected, WNTV 
works better for cartoon image as shown in Fig. \ref{ppreult}.
%Nonlocal TV method fails to recover several texture parts in both cases. While the results of WGL are comparable to those of our method, our method provide an extra parameter to adjust, thus having more possibility to fit different applications.

\subsection{Color Image Inpainting}
Now, we apply the algorithm to color images. The basic settings are similar to the grayscale image examples. In color image, patch becomes a 3D cube. The size we used is 
$11\times11\times3$.  We also use Gaussian weight,
\begin{align*}
\omega(\bx,\by) = \exp\left(-\frac{\|\bx-\by\|^2}{\sigma(\bx)^2}\right)
\end{align*} 
where $\|\bx-\by\|^2$ is the Euclidean distance between patches $\bx$ and $\by$. $\sigma(\bx)$ is the distance between $\bx$ and its 20th nearest neighbor. 
The weight $\omega(\bx,\by)$ is made sparse by setting $\omega(\bx,\by)$ equal to zero if point $\by$ is not among the 50th closest points to point $\bx$. 
The color image is recovered in RGB channels separately. 
\begin{figure}%[H]
\centering
   \subfigure[Original Image.\label{fig:originalimagebarbara}]{\includegraphics[width=3.5cm]{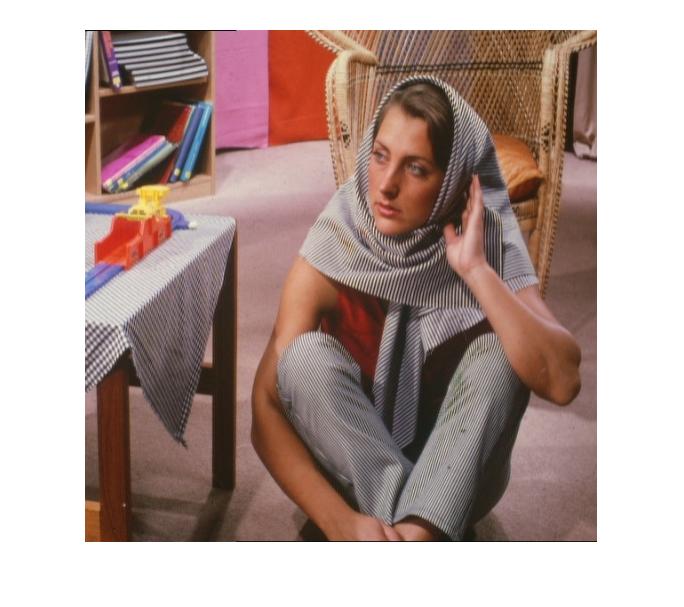}}
   \subfigure[10\% Subsample.\label{fig:sampleimagebarbara}]{\includegraphics[width=3.5cm]{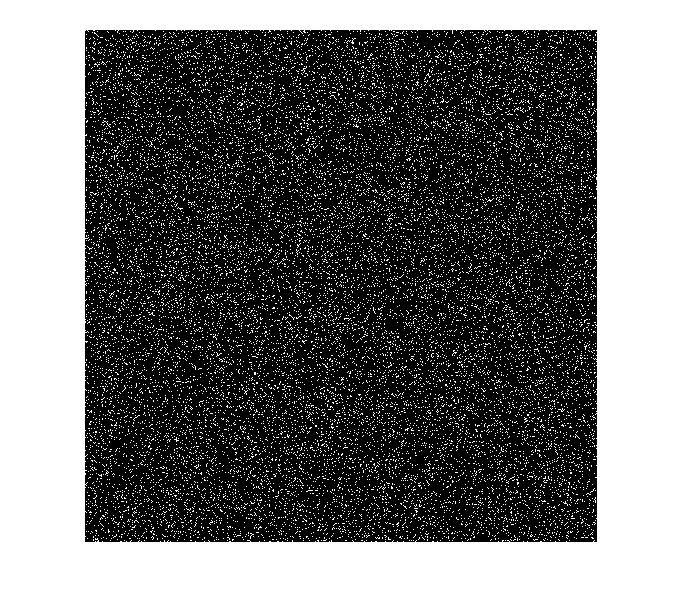}}
   \subfigure[GL (24.31dB).\label{fig:GLbarbara}]{\includegraphics[width=3.5cm]{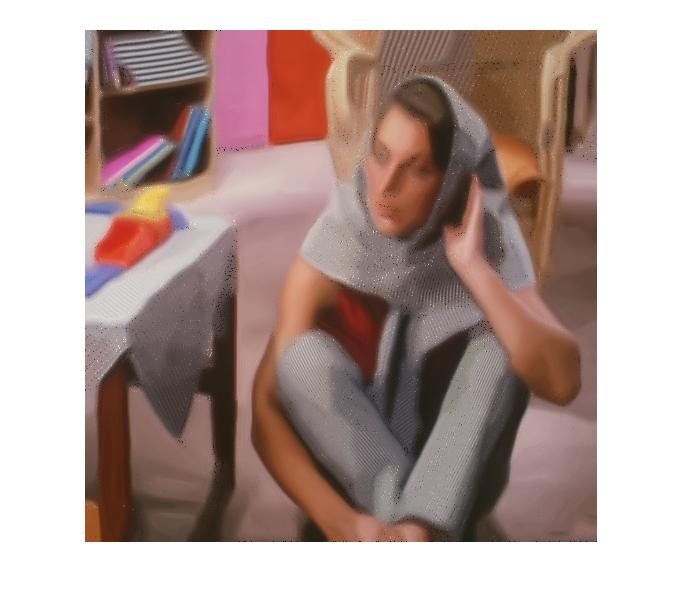}}\\
   \subfigure[NTV (24.38dB).\label{fig:NTVbarbara}]{\includegraphics[width=3.5cm]{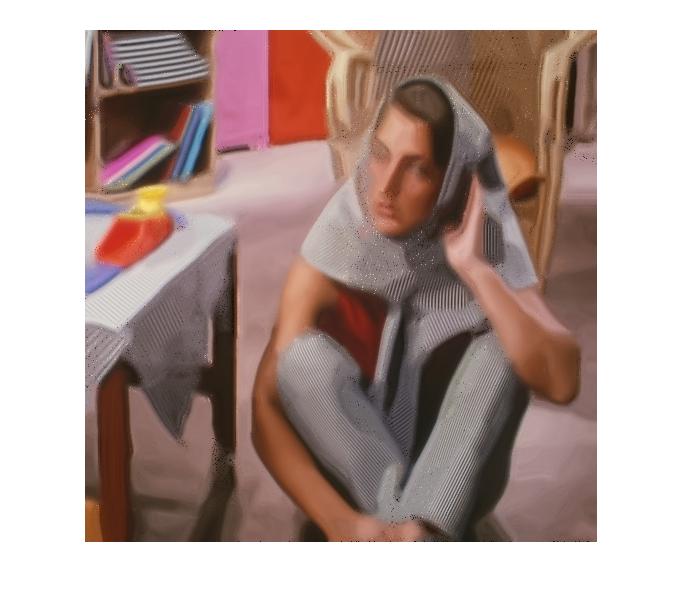}}
   \subfigure[WNLL (26.61dB).\label{fig:WGLbarbara}]{\includegraphics[width=3.5cm]{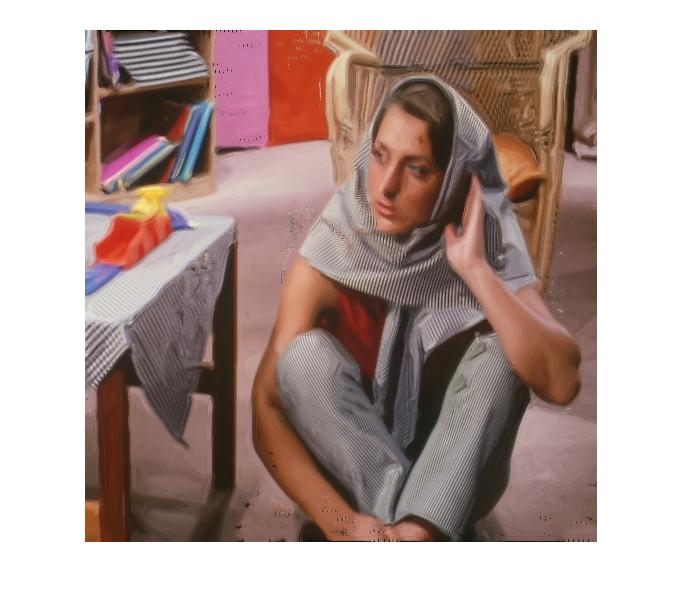}}
   \subfigure[WNTV (26.71dB).\label{fig:WNTVbarbara}]{\includegraphics[width=3.5cm]{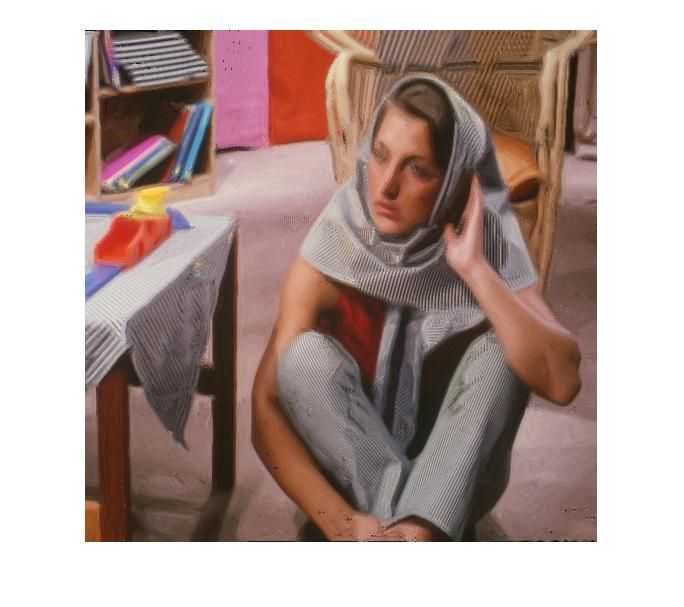}}
   \caption{\label{barbarareult}Results of Graph Laplacian (GL), nonlocal TV (NTV), weighted nonlocal Laplacian (WNLL) and weighted nonlocal TV (WNTV) in the color image of Barbara. % (a): Given image; (b): Subsampled image 10\% sampling; (c): Result of Graph Laplacian, PSNR = 24.31; (d): Result of nonlocal TV, PSNR = 24.38; (e): Result of weighted Graph Laplacian, PSNR = 26.61; (f): Result of weighted nonlocal TV, PSNR = 26.71.
   }
\end{figure}

\begin{figure}%[H]
\centering
   \subfigure[Given Image.\label{fig:originalimagehs}]{\includegraphics[width=4cm]{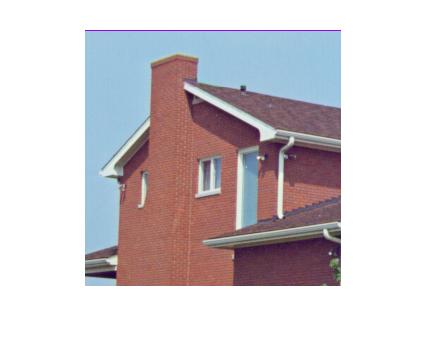}}
   \subfigure[10\% Subsample.\label{fig:sampleimagehs}]{\includegraphics[width=4cm]{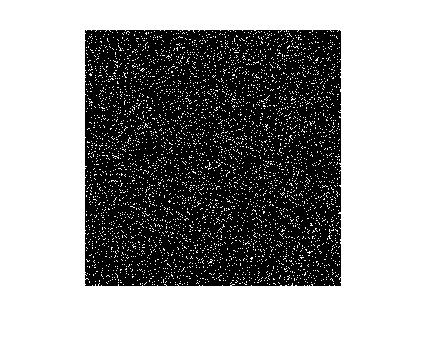}}
   \subfigure[GL (24.28dB).\label{fig:GLhs}]{\includegraphics[width=4cm]{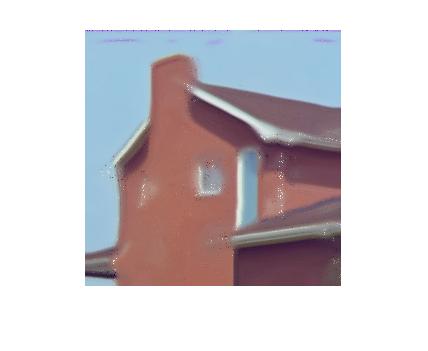}}\\
   \subfigure[NTV (23.81dB).\label{fig:NTVhs}]{\includegraphics[width=4cm]{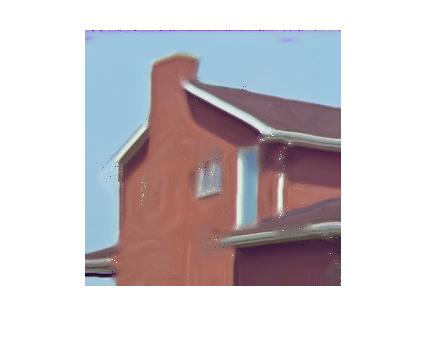}}
   \subfigure[WNLL (26.61dB).\label{fig:WGLhs}]{\includegraphics[width=4cm]{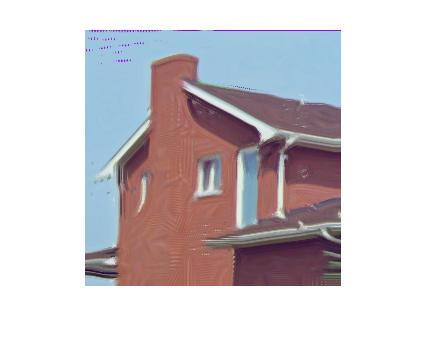}}
   \subfigure[WNTV (27.34dB).\label{fig:WNTVhs}]{\includegraphics[width=4cm]{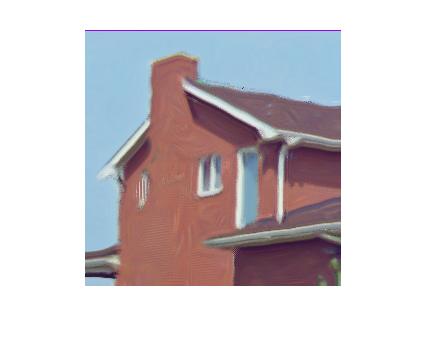}}
   \caption{\label{housereult}Results of Graph Laplacian, nonlocal TV, weighted Graph Laplacian and weighted nonlocal TV applied to color house image. % (a): Given image; (b): Subsampled image 10\% sampling; (c): Result of Graph Laplacian, PSNR = 24.28; (d): Result of nonlocal TV, PSNR = 23.81; (e): Result of weighted Graph Laplacian, PSNR = 26.61; (f): Result of weighted nonlocal TV, PSNR = 27.34.
   }
\end{figure}

We apply our algorithm to Fig. \ref{fig:originalimagebarbara} and \ref{fig:originalimagehs}. Again, WNTV and WNLL outperform NTV and GL. In the image of house, in which 
cartoon dominates, the result of WNTV is better than WNLL. While in the image of Barbara, WNTV and WNLL are comparable since this image is rich in textures. % Obviously, strong artifacts appear near the face in \ref{fig:GLbarbara} and \ref{fig:NTVbarbara}. The same  phenomenon also occurs in \ref{fig:GLhs} and \ref{fig:NTVhs} near the windows. In both cases, we see a much better restoration using WNTV and WGL than standard versions. Also, it is illustrated the proposed method gives a visually smoother restored image. The differece between weighted Graph Laplacian and proposed method seem to be small except some tiny artifacts occur via the precious one.

\section{Image Colorization\label{sec5}}
Colorization is the process of adding color to monochrome images. 
It is usually done by person who is color expert but still this process is time consuming and sometimes could be boring. One way to reduce the working load is only add color in 
part of the pixels by human and using some colorization method to extend the color to other pixels. 

%One usually do image colorization based on some prior knowledge about the object presented in particular images. Here we propose to use weighted nonlocal total variation method for image colorization. The graph constructions and basic settings are the similar to image inpainting application. 

This problem can be natrually formulated as an interpolation on point cloud. The point cloud is constructed by taking patches from the gray image. On the patches, we have three
functions, $u_R$, $u_G$ and $u_B$ corresponding to three channels of the color image. Then WNTV is used to interpolate $u_R$, $u_G$ and $u_B$ over the whole patch set. 
The weight is computed in the same way as that in image inpainting.
% Image colorization we are doing is to learn from a gray style image $f_g:V\rightarrow \mathbb{R}$ to define a mapping from the vertices of that image to a vector of RGB color channels $f_c:V\rightarrow \mathbb{R}^3$, $f_c(x) = [f_c^1(x), f_c^2(x), f^3_c(x)]^T$. What we know is the mapping from a small subset of vertices to the RGB color channels $f_s:S\rightarrow \mathbb{R}^3$, where $S$ is a subset of $V$. 
% \begin{align*}
% f_c(x) = \left\{ \begin{array}{lll}
%     f_s(x), &  & \text{if} \ \ \ x \in V\setminus S,\\
%     0, &  & \text{otherwise}.
%   \end{array} \right.
% \end{align*}
% We then applies each channels of the function $f_c$ to Algorithm \ref{SBWTV}. The patch sets are constructed on the gray style image $f_g$, so does the Gaussian weight function:
% \begin{align*}
% \omega(x,y) = exp(-\frac{||x-y||^2}{\sigma(x)^2})
% \end{align*} 
% where $||x-y||^2$ is the 2-norm of the distance between two feature vectors of $x$ and $y$ on gray image. $\sigma(x)$ is the distance between $x$ and its 20th nearest neighbor. 

\begin{figure}%[H]
\centering
   \subfigure[Original Color Image.\label{fig:originalimagebb}]{\includegraphics[width=3.5cm]{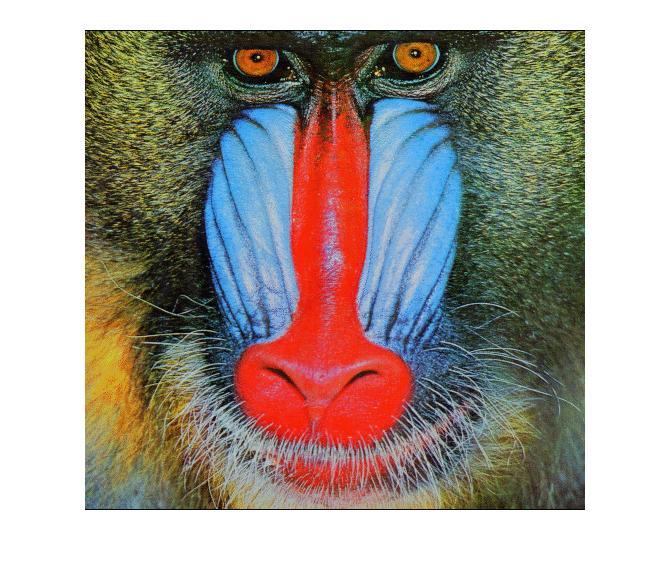}}
%   \subfigure[Subsampled Image.\label{fig:sampleimagebb}]{\includegraphics[width=3.5cm]{images/bb/sampleimagebb.jpg}}
   \subfigure[Gray Style Image.\label{fig:graystylebb}]{\includegraphics[width=3.5cm]{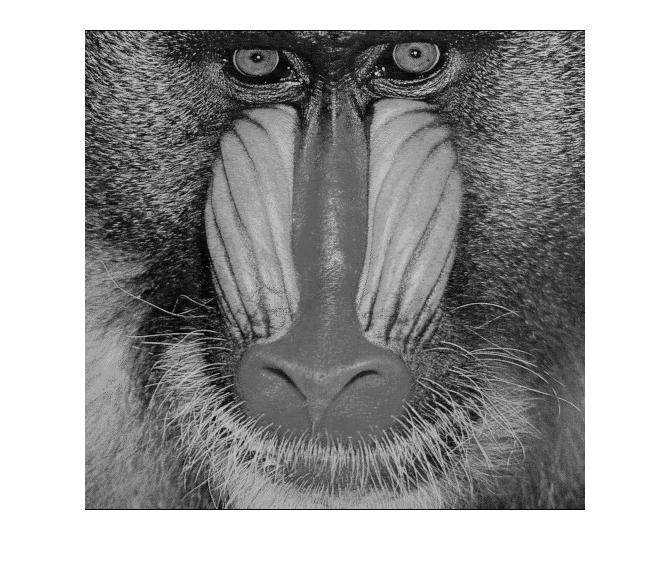}}
   \subfigure[GL (16.46dB).\label{fig:GLcolorbb}]{\includegraphics[width=3.5cm]{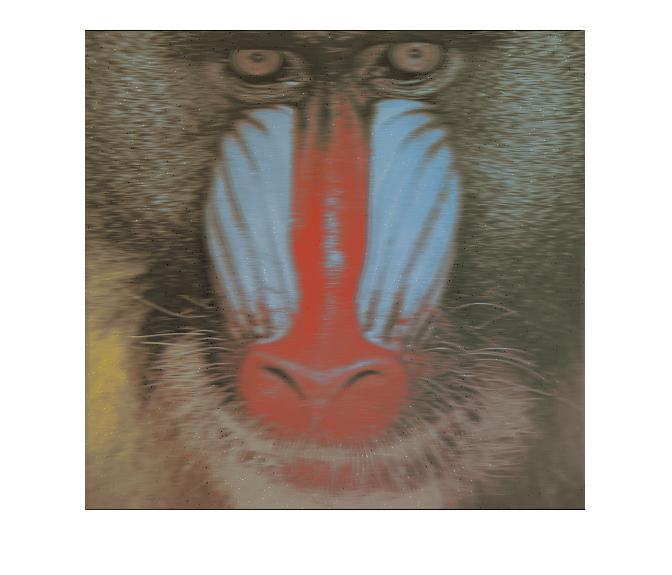}}\\
   \subfigure[NTV (16.20dB).\label{fig:NTVcolorbb}]{\includegraphics[width=3.5cm]{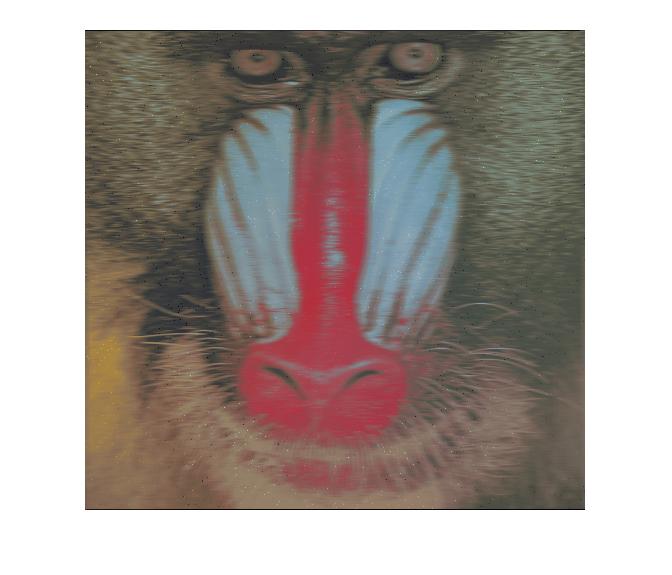}}
   \subfigure[WNLL (20.32dB).\label{fig:sWGLcolorbb}]{\includegraphics[width=3.5cm]{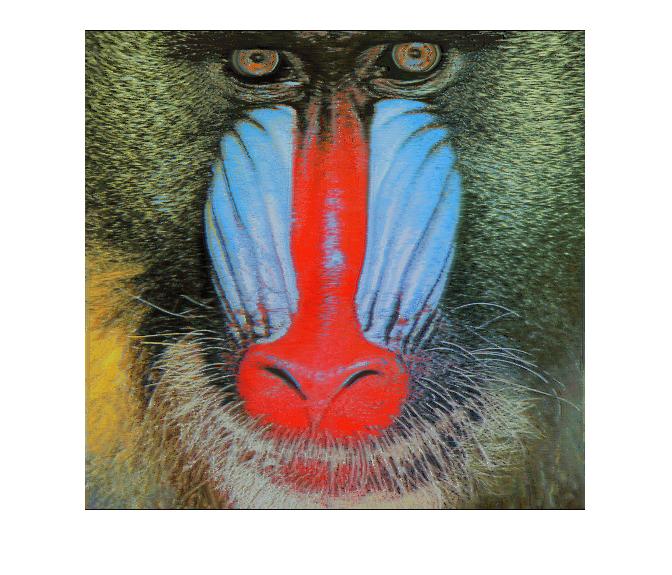}}
    \subfigure[WNTV (20.55dB).\label{fig:WNTVcolorbb}]{\includegraphics[width=3.5cm]{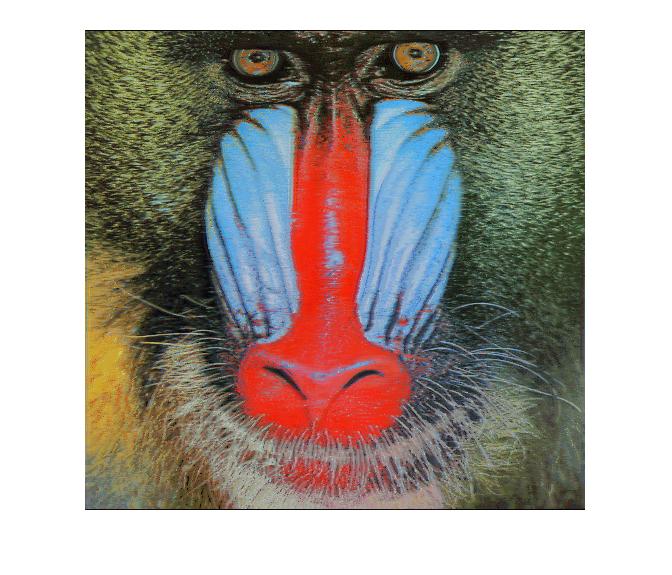}}
   \caption{\label{bbreult}Results of Graph Laplacian (GL), nonlocal TV (NTV), weighted nonlocal Laplacian (WNLL) and weighted nonlocal TV (WNTV) in the baboon image colorization from 1\% samples. % (a): Color image; (b): Subsampled image 1\% sampling; (c): Gray level image; (d): Result of Graph Laplacian, PSNR = 16.46; (e): Result of nonlocal TV, PSNR = 16.20; (f): Result of nonlocal Graph Laplacian, PSNR = 20.32;  (g): Result of weighted nonlocal TV, PSNR = 20.55.
   }
\end{figure}

\begin{figure}%[H]
\centering
   \subfigure[Original Color Image.\label{fig:originalimagef16}]{\includegraphics[width=3.5cm]{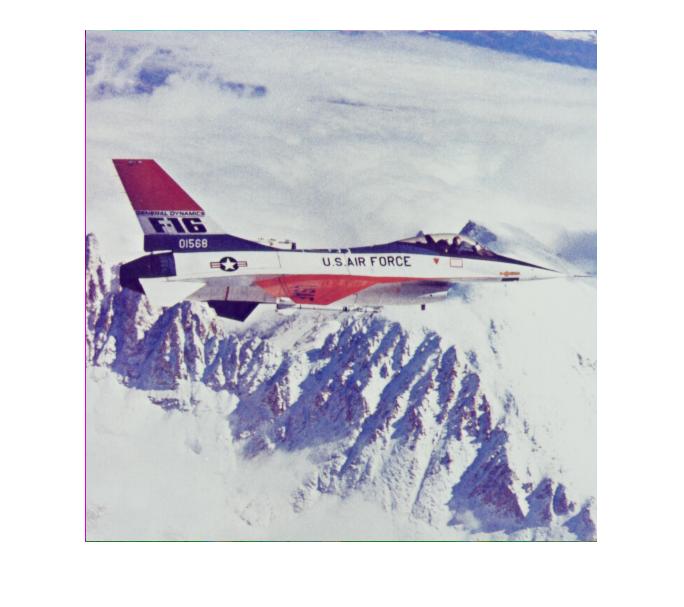}}
%   \subfigure[Subsampled Image.\label{fig:sampleimagef16}]{\includegraphics[width=3.5cm]{images/f16/sampleimagef16.jpg}}
   \subfigure[Gray Style Image.\label{fig:graystylefruit}]{\includegraphics[width=3.5cm]{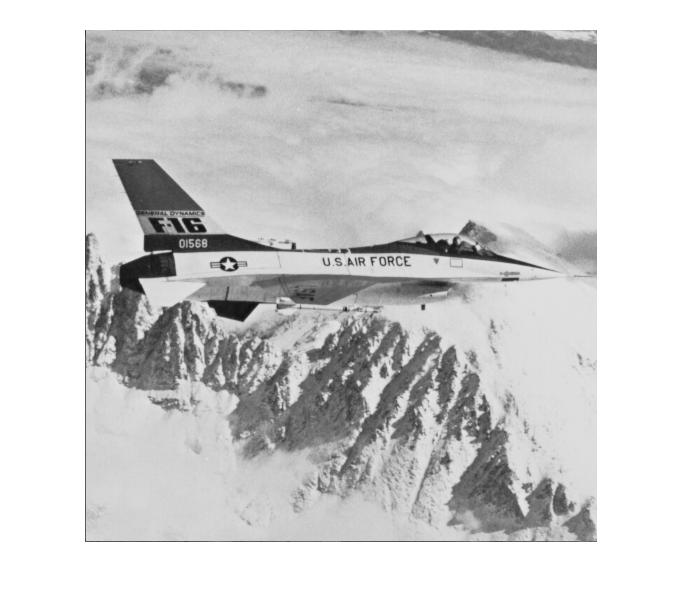}}
   \subfigure[GL (24.87dB).\label{fig:GLcolorfruit}]{\includegraphics[width=3.5cm]{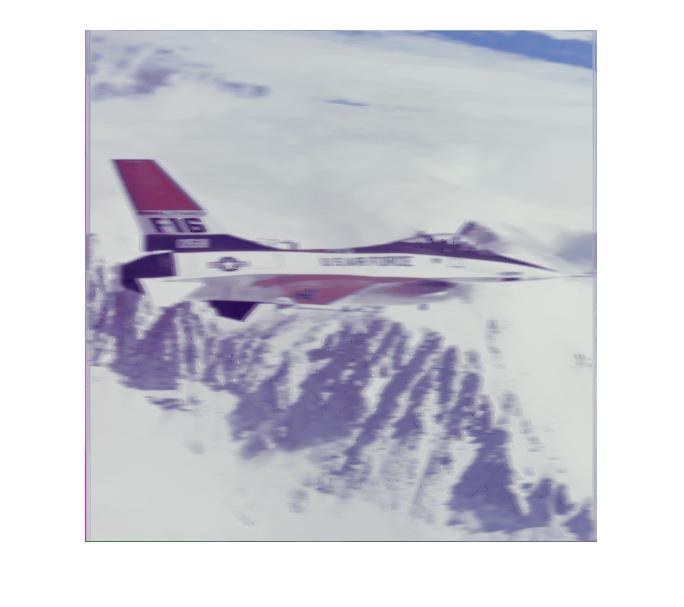}}\\
   \subfigure[NTV (24.75dB).\label{fig:NTVcolorfruit}]{\includegraphics[width=3.5cm]{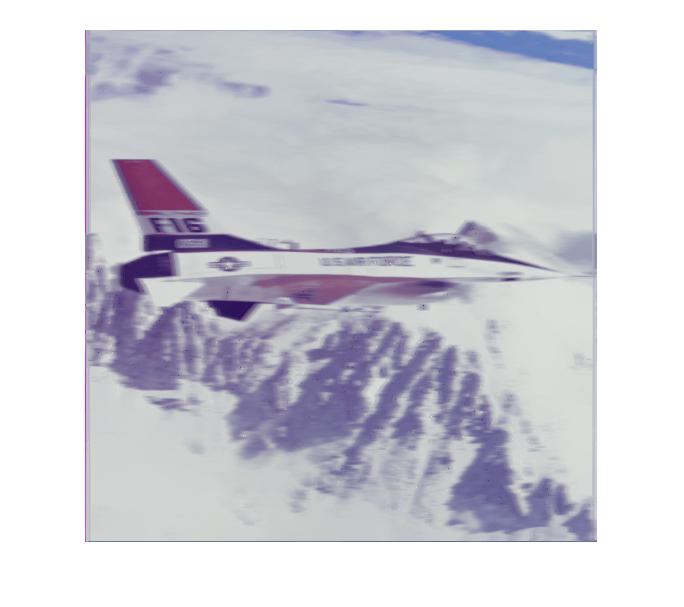}}
   \subfigure[WNLL (28.57dB).\label{fig:sWGLcolorfruit}]{\includegraphics[width=3.5cm]{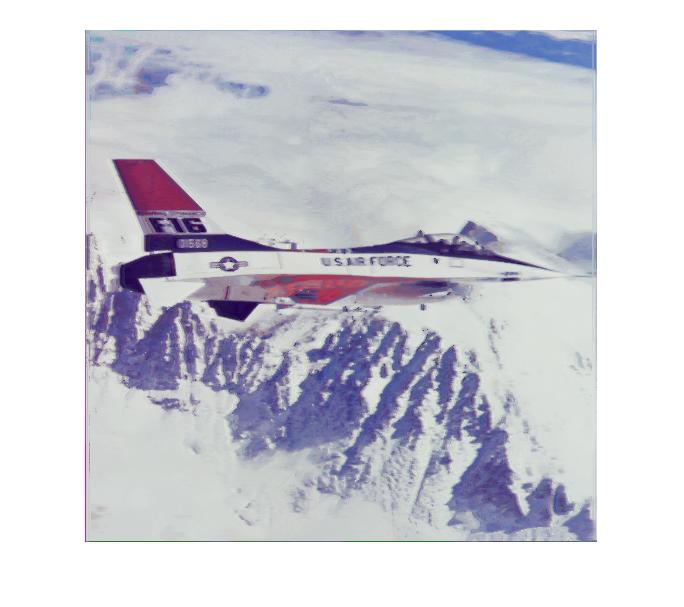}}
    \subfigure[WNTV (28.94dB).\label{fig:WNTVcolorfruit}]{\includegraphics[width=3.5cm]{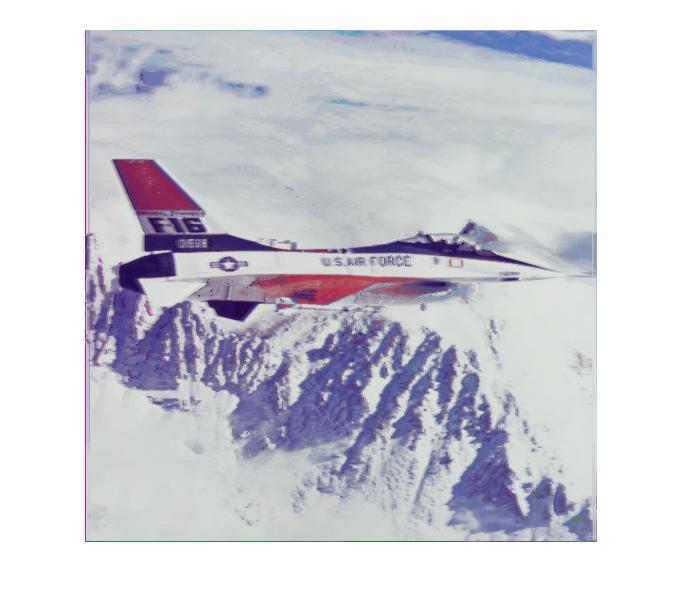}}
   \caption{\label{f16reult}Results of Graph Laplacian (GL), nonlocal TV (NTV), weighted nonlocal Laplacian (WNLL) and weighted nonlocal TV (WNTV) in the F-16 image colorization from 1\% samples. % (a): Color image; (b): Subsampled image 1\% sampling; (c): Gray level image; (d): Result of Graph Laplacian, PSNR = 24.87; (e): Result of nonlocal TV, PSNR = 24.75; (f): Result of nonlocal Graph Laplacian, PSNR = 28.57;  (g): Result of weighted nonlocal TV, PSNR = 28.94.
   }
\end{figure}

\begin{figure}%[H]
\centering
   \subfigure[Original Color Image.\label{fig:originalimagebff}]  {\includegraphics[width=4cm]{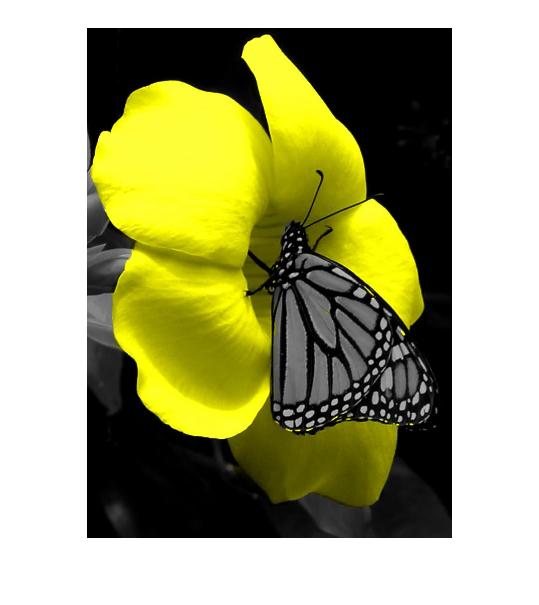}}
  % \subfigure[Subsampled Image.\label{fig:sampleimagebff}]{\includegraphics[width=4cm]{images/bff/sampleimagebff.jpg}}
   \subfigure[Gray Style Image.\label{fig:graystylebff}]{ \includegraphics[width=4cm]{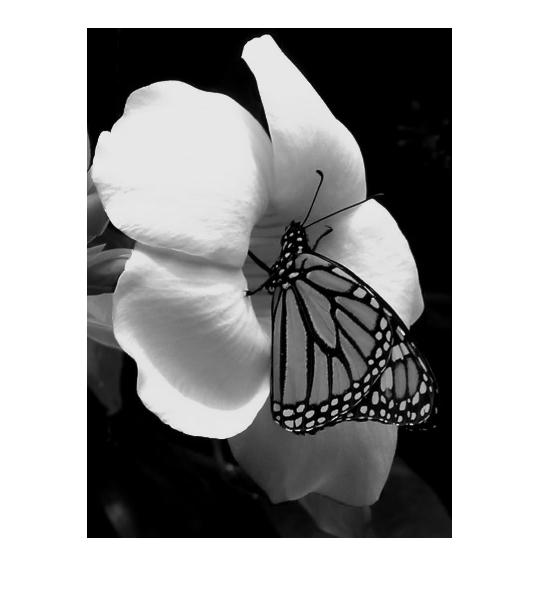}}
   \subfigure[GL (28.30dB).\label{fig:GLcolorbff}]{ \includegraphics[width=4cm]{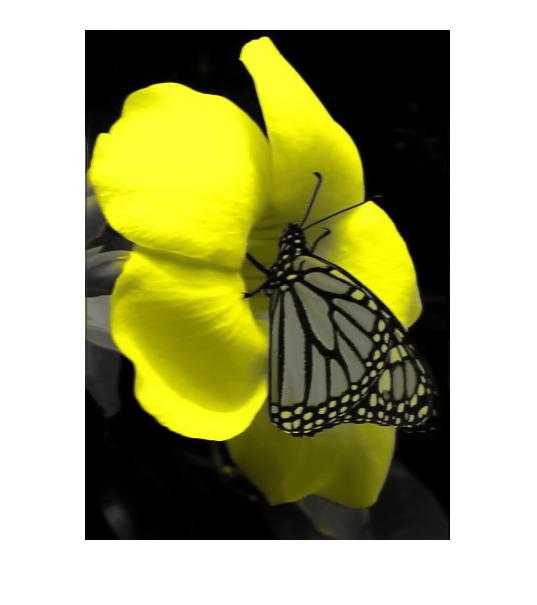}}\\
   \subfigure[NTV (28.43dB).\label{fig:NTVcolorbff}]{ \includegraphics[width=4cm]{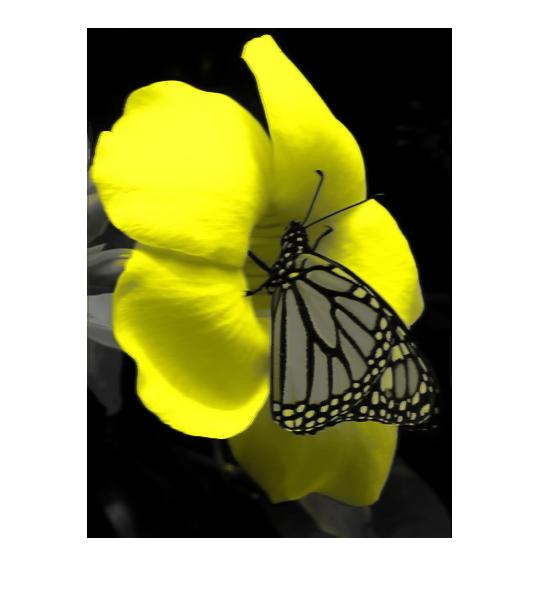}}
   \subfigure[WNLL (31.98dB).\label{fig:WGLcolorbff}]{ \includegraphics[width=4cm]{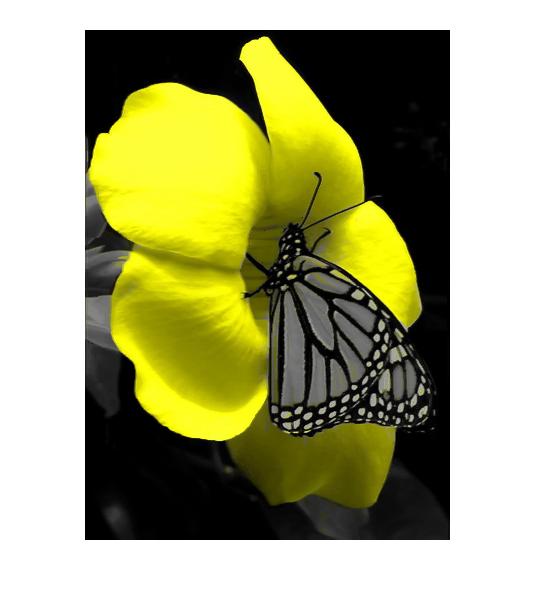}}
   \subfigure[WNTV (32.35dB).\label{fig:WNTVcolorbff}]{ \includegraphics[width=4cm]{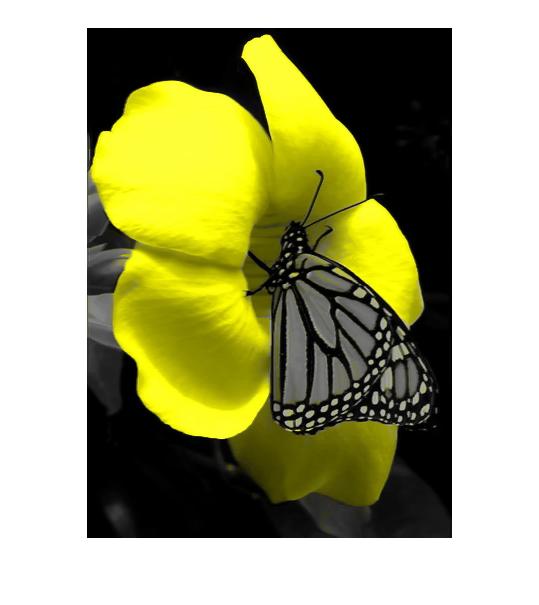}}
   \caption{\label{bffreult} Results of Graph Laplacian (GL), nonlocal TV (NTV), weighted nonlocal Laplacian (WNLL) and weighted nonlocal TV (WNTV) in the butterflyflower image colorization from 1\% samples. % (a): Color image; (b): Subsampled image 1\% sampling; (c): Gray level image; (d): Result of Graph Laplacian, PSNR = 28.30; (e): Result of nonlocal TV, PSNR = 28.43; (f): Result of weighted Graph Laplacian, PSNR = 31.98; (g): Result of weighted nonlocal TV, PSNR = 32.35.
   }
\end{figure}

The colorization results from 1\% samples are demonstrated in Fig. \ref{bbreult}, \ref{f16reult} and \ref{bffreult}. Face of the baboon, snow mountains and wings of butterfly are not properly colored in GL and NTV. The face of baboon are blured. Part of the wings are colored in yellow by mistake. Snow mountain and text on F16 are also blured. In WNLL and WNTV, they are all properly colored. In addition, PSNR value also suggest that 
WNTV has the best performance.

\section{Conclusion\label{sec6}}

In this paper, we propose a weighted nonlocal total variation (WNTV) model for interpolations on high dimensional point cloud. This model can be solved efficiently 
by split Bregman iteration. Numerical tests in semi-supervised learning, image inpainting and image colorization demonstrate that WNTV is an effective and efficient method in 
image processing and data analysis.
\section*{Reference}
\bibliographystyle{abbrv}%{elsarticle-harv}
\bibliography{reference.bib}

\end{document}